\documentclass{applemlr}

\usepackage{amsmath}
\usepackage{enumerate}
\usepackage{algorithm}
\usepackage{algpseudocode}
\usepackage{amsfonts}
\usepackage{amsthm}
\usepackage{cleveref}
\usepackage{diagbox}
\usepackage{colortbl}
\usepackage{amssymb}
\usepackage{xspace}
\usepackage{wrapfig}
\usepackage{adjustbox}
\usepackage{tabularx}
\usepackage{booktabs}
\usepackage{mathtools}
\usepackage{tikz}
\usepackage{enumitem}
\usepackage{silence}
\usepackage{dsfont}
\usepackage[table]{xcolor}
\usepackage[dvipsnames]{xcolor}
\usepackage{multirow}
\usepackage{makecell}
\usepackage{xfakebold}

\usepackage{amsmath,amsfonts,bm}

\def\eqref#1{equation~\ref{#1}}

\def\1{\bm{1}}

\DeclareMathAlphabet{\mathsfit}{\encodingdefault}{\sfdefault}{m}{sl}
\SetMathAlphabet{\mathsfit}{bold}{\encodingdefault}{\sfdefault}{bx}{n}

\DeclareMathOperator*{\argmax}{arg\,max}

\definecolor{textgray}{HTML}{6E6E73}
\usetikzlibrary{positioning, calc}
\usetikzlibrary{decorations.pathmorphing}

\makeatletter
\patchcmd{\wrong@fontshape}{\@gobbletwo}{}{}{}
\makeatother
\numberwithin{equation}{section}
\makeatletter
\AtBeginDocument{
  \urlstyle{sf}
  
}
\makeatother

\definecolor{light}{RGB}{125, 125, 125}
\crefname{tcb@cnt@pbox}{code}{code}
\Crefname{tcb@cnt@pbox}{Code}{Code}
\crefname{assumption}{assumption}{assumption}
\Crefname{assumption}{Assumption}{Assumptions}

\newtcolorbox[auto counter]{pbox}[2][]{
  colback=white,
  title=Code~\thetcbcounter: #2,
  #1,fonttitle=\sffamily,
  fontupper=\sffamily,
  arc=2pt,
  colframe=bgcolor,
  coltitle=fgcolor,
  colbacktitle=bgcolor,
  toptitle=0.25cm,
  bottomtitle=0.125cm
}

\makeatletter
\newcommand\applefootnote[1]{%
  \begingroup
  \renewcommand\thefootnote{}%
  \renewcommand\@makefntext[1]{\noindent##1}%
  \footnote{#1}%
  \addtocounter{footnote}{-1}%
  \endgroup
}
\makeatother

\definecolor{cverbbg}{gray}{0.90}

\title{Revisiting ASR Error Correction with Specialized Models}

\author{Zijin Gu}
\author{Tatiana Likhomanenko}
\author{Richard He Bai}
\author{Erik McDermott}
\author{Ronan Collobert}
\author[\dag]{Navdeep Jaitly}

\affiliation{Apple}
\affiliation[\dag]{Google; work done at Apple}

\abstract{
Language models play a central role in automatic speech recognition (ASR), yet most methods rely on text-only models unaware of ASR error patterns.
Recently, large language models (LLMs) have been applied to ASR correction, but introduce latency and hallucination concerns.
We revisit ASR error correction with compact seq2seq models, trained on ASR errors from real and synthetic audio.
To scale training, we construct synthetic corpora via cascaded TTS and ASR, finding that matching the diversity of realistic error distributions is key.
We propose correction-first decoding, where the correction model generates candidates rescored using ASR acoustic scores.
With 15x fewer parameters than LLMs, our model achieves 1.5/3.3\% WER on
LibriSpeech test-clean/other, outperforms LLMs, generalizes across ASR
architectures (CTC, Seq2seq, Transducer) and diverse domains, and provides
precise corrections in the low-error regime where LLMs struggle.}

\metadata[Correspondence]{\sffamily Zijin Gu: \url{zijin@apple.com}}
\date{\sffamily\today}

\begin{document}

\maketitle

\section{Introduction}

Language models are a key component of modern automatic speech recognition (ASR) systems, providing linguistic constraints that complement acoustic modeling~\cite{synnaeve2019end}.
In conventional pipelines, this interaction is typically realized through shallow fusion, rescoring, or related heuristics that combine acoustic scores with probabilities from text-only language models~\cite{chorowski2016towards,kannan2017analysis}.
While effective, these approaches rely on language models that are trained independently of ASR systems and are therefore unaware of the error patterns produced during decoding, often requiring careful tuning and increased computational cost.

Recent advances in large language models (LLMs) have renewed interest in
post-hoc ASR
correction~\cite{chen2023hyporadise,yang2023generative,hu2024large,fathullah2024prompting,radhakrishnan2023whispering,
fang2025fewer, liu2025listening}, demonstrating that strong generative models can fix recognition errors using
powerful linguistic priors.
However, LLM-based correction introduces practical challenges, including high
inference latency, limited controllability, and a tendency to hallucinate
corrections that diverge from the acoustic evidence~\cite{hu2024large, fang2025fewer, liu2025listening}.
Moreover, LLMs are fundamentally ill-suited for correcting \textit{already-accurate} ASR systems: reducing WER from 30\% to 20\% requires coarse linguistic fixes that LLMs can provide, but reducing WER from 5\% to 3\% demands precise, character-level corrections that preserve acoustic evidence---a regime where LLMs struggle and often degrade performance through over-correction~\cite{wu2023decoder,yang2023generative}.
This raises a fundamental question: \textit{do specialized error correction models provide a better accuracy--efficiency trade-off than generic large language models for ASR error correction?}

In this work, we revisit ASR error correction with specialized language models.
Rather than relying on language models trained only on clean text, which are agnostic to ASR error patterns, or on generic LLMs for post-hoc correction, we propose to explicitly learn the conditional distribution that maps corrupted ASR hypotheses to clean transcripts.
We study error correction language models (ECLMs), compact conditional sequence-to-sequence models trained specifically for this purpose.
This separation offers a key advantage: the acoustic model can focus purely on mapping speech to plausible token sequences, while the ECLM learns to correct the systematic errors that acoustic models make---without requiring joint optimization or heuristic score combination.

A central challenge in training such models is the scarcity of paired ASR hypotheses and reference transcripts.
To overcome this limitation, we construct large-scale training data by cascading text-to-speech (TTS) systems with ASR models~\cite{guo2019spelling,hu2021synt}, supplemented with real ASR hypotheses from labeled audio, enabling the generation of arbitrarily large noisy-clean text pairs that capture realistic error distributions.
Through systematic ablations, we find that the effectiveness of ECLMs depends less on the perceptual quality of synthetic speech and more on matching the diversity and structure of realistic ASR error distributions.
We achieve this through multi-speaker synthesis, noise augmentation, and mixing synthetic and real data sources.
Crucially, this decoupled approach scales efficiently: text corpora can be expanded independently of acoustic data, and a single ECLM transfers across different ASR architectures without retraining.

Beyond model training, we also revisit decoding.
In conventional approaches, ASR generates an $n$-best list using beam search, e.g., with an $n$-gram LM, and a neural LM then rescores these hypotheses.
Rather than rescoring ASR hypotheses with a language model, we propose \textit{correction-first decoding}, in which the ECLM generates candidate corrections that are subsequently evaluated using acoustic scores from the ASR system.
This approach enables interaction between acoustic and linguistic evidence without relying on expensive ASR beam sizes and language-model rescoring.

Our main contributions are as follows:
\begin{itemize}
    \item We demonstrate that our ECLM achieves state-of-the-art WER on LibriSpeech (1.5\% test-clean, 3.3\% test-other) without using external audio data, matching self-supervised approaches that use 60k hours of unlabeled audio, while outperforming both off-the-shelf and fine-tuned LLMs with 15$\times$ fewer parameters, lower latency, and no hallucination.
    \item We propose correction-first decoding, which outperforms conventional neural LM rescoring while offering comparable computational cost.
    \item We demonstrate that ECLMs generalize across ASR architectures (QuartzNet, Conformer, Whisper, Parakeet-TDT) spanning CTC, seq2seq, and transducer paradigms, and across diverse domains, reducing WER by 13\% relative on average across eight datasets spanning conversational and prepared speech.
    \item Through systematic ablations, we identify key ingredients for effective training data: multi-speaker TTS, noise augmentation, mixing synthetic and real data, and---counterintuitively---noisier TTS systems outperform high-quality ones.
\end{itemize}

Our results show that in the low-error regime where LLMs overcorrect and hallucinate, compact specialized models that learn the ASR error distribution offer a superior alternative to both neural LM rescoring and generic LLMs.

\section{Related work}

\noindent\textbf{LM integration for ASR}~~~Finding better techniques to leverage LMs most effectively with acoustic models has been a long standing research problem~\cite{vesely13_interspeech,toshniwal2018comparison}.
Earlier techniques back-propagated errors from LMs into ASRs using sequence discriminative criterion (e.g. Maximum Mutual Information)~\cite{kingsbury2009lattice,jaitly2012application,vesely13_interspeech}.
Later approaches attempted to merge features of text-only LMs with features of ASRs in either the shallower or deeper layers of the model~\cite{gulcehre2015using,sriram2017cold,kannan2017analysis}.
Another line of work~\cite{ilme,McDermott2019ADR,hat} attempts to subtract the language model learned in the end-to-end ASR models, while integrating external LMs.

\noindent\textbf{Error correction models}~~~Error correction models post-process outputs from an ASR system to fix errors~\cite{guo2019spelling,shivakumar2019learning,tanaka2018neural,zhang19g_interspeech}.
These approaches typically convert first-pass ASR hypotheses into cleaned up text.
Even with just one best hypothesis from the ASR model, Transformer-based error correction models have shown improvements over the baseline acoustic models and acoustic models with $n$-gram language models~\cite{hrinchuk2020correction, zhao2021bart}. Further improvements can be made by using $n$-best ASR hypotheses as inputs to the clean up model as this provides more information~\cite{leng2021fastcorrect, ma2023n}; by using more advanced WER based metrics, instead of cross-entropy for model optimization~\cite{MWELSTM}; or by using different variants that incorporate more compact inputs, such as phonemes~\cite{wang2020asr,dutta2022error} and word lattices~\cite{ma2020neural,dai2022latticebart}.
However, the efficacy of these approaches has been limited by the scarcity of paired ASR output and ground truth transcriptions, \textit{and thus have not been able to outperform conventional neural LM integration.}
As a result most approaches start with a pretrained language model, e.g., BART~\cite{lewis2019bart}, that is finetuned with the limited noisy ASR data~\cite{zhao2021bart, dutta2022error,ma2023n}.
Others incorporate different data augmentation strategies, such as SpecAugment~\cite{ma2023n} and dropout~\cite{hrinchuk2020correction}.
While these are helpful, the improvement may be limited by the data size.
\cite{guo2019spelling} have attempted to resolve this problem by using synthetic data for training error correction models on top of listen, attend and spell (LAS)~\cite{chan2015listen} models. They show that error correction can improve results when combined with neural LMs, but error correction by itself is inferior to neural LM integration.

\noindent\textbf{LLM-based error correction}~~~More recently, large language models (LLMs) have been explored for ASR error correction with promising results.
HyPoradise~\cite{chen2023hyporadise} proposed using LLMs with hypotheses-to-transcription prompting, while~\cite{yang2023generative} showed that LLMs can correct ASR errors through in-context learning and task-activating prompts.
\cite{hu2024large} demonstrated that LLMs are efficient learners for noise-robust speech recognition with minimal finetuning.
Other works have explored prompting LLMs with speech features directly~\cite{fathullah2024prompting,wu2023decoder} or combining Whisper with LLaMA for cross-modal correction~\cite{radhakrishnan2023whispering}.
While LLM-based methods show promise, they have notable limitations for practical deployment:
(i) \textit{Latency}: API-based LLMs introduce network latency unsuitable for real-time ASR;
(ii) \textit{Cost}: per-token pricing makes large-scale transcription expensive;
(iii) \textit{Reliability}: proprietary models may change or be discontinued, creating dependency risks and making results hard to reproduce;
(iv) \textit{Offline deployment}: many applications require on-device processing without internet access.
In contrast, our ECLM is a compact ($<1B$ parameters), self-contained model that can be deployed offline while being specifically trained for error correction.
More fundamentally, LLM-based methods excel at correcting \textit{linguistic} errors (grammar, word choice, semantic coherence) but struggle with \textit{acoustic} errors (phonetic confusions, homophones, character-level mistakes) because they lack knowledge of the ASR's error distribution.
Our ECLM is trained on TTS-generated errors that mimic the ASR's actual mistake patterns, enabling fine-grained character-level corrections that generic LLMs cannot achieve.

\begin{figure}[t]
\centering
\includegraphics[width=0.6\columnwidth]{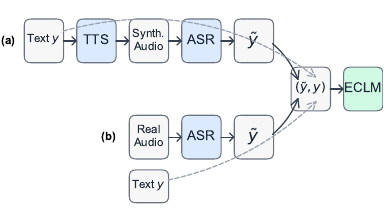}
\caption{Data generation pipeline for training the error correction model. Noisy hypotheses $\tilde{y}$ are obtained by running ASR on either (a)~synthetic audio generated via multi-speaker TTS from a text corpus, or (b)~real audio. These are paired with clean references $y$ to create training examples $(\tilde{y}, y)$ for the ECLM.}
\label{fig:data_pipeline}
\end{figure}

\section{Error correction for speech recognition}\label{sec:methods}

\subsection{Probabilistic formulation}

To motivate the error correction approach and our variant of it, we consider a probabilistic model $p_{\tt{corr}}(y|x)$, which we call correction-based speech recognition. This is a cascade of two stochastic, discrete models -- an ASR model which produces sequences~$\tilde{y}$ of discrete tokens from audio input $x$ with probability $p_{\tt{ASR}}(\tilde{y}|x)$, and an ECLM which transforms an input sequence, $\tilde{y}$, into an output sequence $y$ with probability $p_{\tt{ECLM}}(y|\tilde{y})$. Here $\tilde{y}$ denotes any ASR hypothesis and $\hat{y} = \argmax_{\tilde{y}} p_{\tt{ASR}}(\tilde{y}|x)$ denotes the greedy hypothesis. Under this model
\begin{equation}\label{eqn:DSR}
\log p_{\tt{corr}}(y|x) = \log \left(\sum_{\tilde{y}}p_{\tt{ECLM}}(y|\tilde{y}) p_{\tt{ASR}}(\tilde{y}|x)\right).
\end{equation}
Most error correction approaches optimize
\begin{equation}\label{eqn:optimization_eqn}
\log p_{\tt{ECLM}}(y|\hat{y}), \,\,\,
\text{where } \hat{y} = \argmax_{\tilde{y}} p_{\tt{ASR}}(\tilde{y}|x),
\end{equation}
which can be viewed as an approximation to the lower bound of Equation~(\ref{eqn:DSR})
\begin{equation}
\log p_{\tt{corr}}(y|x) \geq \sum_{\tilde{y}} p_{\tt{ASR}}(\tilde{y}|x) \log p_{\tt{ECLM}}(y|\tilde{y})
\end{equation}
under the assumption that the ASR model is sharply peaked on a single transcript (the inequality comes from applying Jensen's inequality). Note, however, that using samples from the posterior of the model, $p_{\tt{corr}}(\tilde{y}|x, y)$, would be the correct distribution to optimize the above model on, but we (and prior works on error correction) do not follow that approach here.

\subsection{Data generation}
We would like to optimize Equation~(\ref{eqn:optimization_eqn}) in order to improve $\log p_{\tt{corr}}$.
This requires paired data $\{(\hat{y}, y)\}$ where $\hat{y}$ is a corrupted ASR hypothesis and $y$ is the clean reference.
Unlike conventional LMs which model clean text alone, ECLMs are conditional models whose input distribution must reflect realistic ASR errors.
The central challenge is therefore constructing a training distribution that is close to the true error distribution encountered at test time.
We address this through three complementary strategies: synthetic data generation via TTS, mixing in real ASR hypotheses, and noise augmentation.

\noindent\textbf{Synthetic data via TTS}~~~Given a large text corpus with distribution $p(y)$, we generate paired data by cascading a TTS system with an ASR system (see Figure~\ref{fig:data_pipeline}(a)).
For each sentence $y$, the TTS system synthesizes audio $\tilde{x}$, and the ASR system produces a hypothesis $\hat{y}$ from $\tilde{x}$.
The resulting dataset $\{(\hat{y}, y)\}$ is used to train the ECLM.
This approach captures the biases of the ASR system---phonetic confusions, insertion/deletion patterns, and character-level errors---more faithfully than random text corruptions~\cite{raml}.\footnote{As initial experiments we also tried carefully designed by hand corruptions, e.g. $n$-gram modifications of the text which did not work in practice.}

\noindent\textbf{Mixing real and synthetic data}~~~While TTS-based generation can scale to arbitrarily large text corpora, the resulting error distribution may not perfectly match that of real speech.
We therefore mix a small proportion of real ASR hypotheses (obtained by running the ASR on labeled audio) into the synthetic training data.
This grounds the ECLM's error distribution more closely to real conditions and provides complementary error patterns that TTS alone may not produce.

\noindent\textbf{Noise augmentation}~~~To further diversify the error distribution, we apply two forms of augmentation: (1)~random character substitution in the input transcript, and (2)~frequency masking of the spectrogram before ASR inference.
Both strategies increase the variety of errors without shifting the distribution away from realistic ASR mistakes.

We ablate all of these factors---TTS system choice, real data mixing, noise augmentation, and scaling---in Sections~\ref{sec:noise_effect}--\ref{sec:scaling}.

\subsection{Decoding techniques}\label{sec:denoising_techniques}
At inference time, given audio $x$, we seek the transcript $y^*$ that maximizes
\begin{equation}
y^* = \argmax_y \sum_{\tilde{y}}p_{\tt{ECLM}}(y|\tilde{y}) p_{\tt{ASR}}(\tilde{y}|x).
\end{equation}
Since the exact optimization is intractable, we consider two approximations: \textit{greedy decoding}, which requires no access to acoustic scores, and \textit{correction-first decoding}, which reintroduces acoustic evidence through rescoring.

\noindent\textbf{Greedy decoding}~~~The ECLM takes the greedy ASR hypothesis $\hat{y}$ and decodes $y^*$ independently of acoustic scores\footnote{For CTC models, $\hat{y}$ is obtained by taking the argmax over label probabilities at each frame, deduplicating, and removing blanks.}
\begin{equation}
y^* = \argmax_y p_{\tt{ECLM}}(y|\hat{y}); \,\, \hat{y} = \argmax_{\tilde{y}} p_{\tt{ASR}}(\tilde{y}|x),
\end{equation}
where both maximizations are approximated by greedy search.
This is the simplest form of error correction: it operates on text alone, without access to audio or ASR scores---a property not achievable with conventional LM rescoring.

\begin{algorithm}[h]
\caption{Correction-first decoding}\label{alg:correction_first}
\begin{algorithmic}[1]
\Require Audio $x$, ASR model $p_{\tt{ASR}}$, ECLM $p_{\tt{ECLM}}$, beam size~$B$, weight $\lambda$
\Ensure Corrected transcript $y^*$
\State $\hat{y} \leftarrow \argmax_{\tilde{y}} p_{\tt{ASR}}(\tilde{y} | x)$ \hfill \Comment{ASR greedy decoding}
\State $\mathcal{Y} \leftarrow \text{BeamSearch}(p_{\tt{ECLM}}(\cdot|\hat{y}), B)$ \hfill \Comment{ECLM generates $n$-best}
\For{each $y \in \mathcal{Y}$}
    \State Compute $\log p_{\tt{ASR}}(y|x)$ \hfill \Comment{ASR acoustic rescore}
\EndFor
\State $y^* \leftarrow \argmax_{y \in \mathcal{Y}} \big(\lambda \log p_{\tt{ECLM}}(y|\hat{y}) + \log p_{\tt{ASR}}(y|x)\big)$
\end{algorithmic}
\end{algorithm}

\noindent\textbf{Correction-first decoding}~~~To reintroduce acoustic evidence, we propose \textit{correction-first decoding} (Algorithm~\ref{alg:correction_first}).
The ECLM generates an $n$-best list of corrected candidates via beam search from the greedy ASR hypothesis $\hat{y}$, and each candidate is rescored using a weighted combination of the ECLM score and the ASR acoustic score, with a single hyperparameter $\lambda$ tuned on the validation set.
This approach assumes the correct transcript appears within the ECLM's beam; in practice, a beam size of 10 suffices.

\begin{figure}[h]
\centering
\includegraphics[width=0.6\columnwidth]{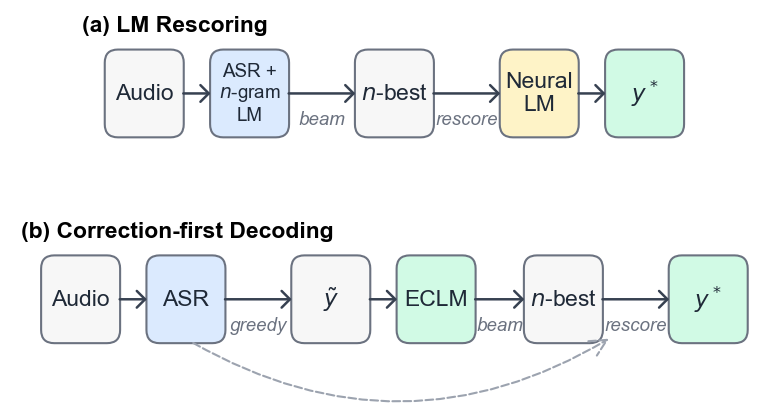}
\caption{Comparison of decoding approaches. (a)~LM rescoring: ASR generates an $n$-best beam, e.g., with $n$-gram LM, then neural LM rescores. (b)~Correction-first decoding: ASR produces a greedy hypothesis, ECLM generates candidates via beam search, and ASR rescores using acoustic likelihood.}
\label{fig:decoding}
\end{figure}

This reverses the conventional neural LM rescoring pipeline (Figure~\ref{fig:decoding}): in standard rescoring~\cite{synnaeve2019end}, the \textit{ASR generates a beam} (typically with a weak $n$-gram LM), which is rescored by the neural LM.\footnote{Joint decoding with ASR and neural LM is also possible but more expensive, and in practice the oracle WER of the first-pass $n$-gram beam is already very low.}
In correction-first decoding, the \textit{ECLM generates the beam} from a single greedy ASR output, and ASR scores are used for rescoring. We adopt nucleus sampling with threshold $0.9$ to ensure full-length beams.
The computational cost of both approaches is comparable, while ECLM greedy decoding is cheaper as it avoids beam search entirely.

\section{Experimental details}\label{sec:exp-details}

\subsection{Training data}\label{sec:tts}
Unless stated otherwise, all ECLMs are trained on data derived from the LibriSpeech~\cite{panayotov2015librispeech} LM corpus (40M sentences, 800M words).
We synthesize audio from this corpus using three TTS systems: Tacotron-2~\cite{shen2018natural} (single-speaker), YourTTS~\cite{casanova2023yourtts} (zero-shot multi-speaker), and RichTTS~\cite{bai2024dmel} (zero-shot multi-speaker with D-vector speaker embeddings~\cite{dvectors}).
For multi-speaker systems, we randomly select speakers from the LibriSpeech ASR training set.
The synthesized audio is processed by a CTC-based ASR system~\cite{graves2006connectionist} to produce hypotheses, yielding paired data $\{(\hat{y}, y)\}$ for ECLM training.
We compare TTS systems in Section~\ref{sec:tts_analysis}.

As discussed in Section~\ref{sec:methods}, we also incorporate a small proportion (10\%) of real ASR hypotheses obtained by running the baseline ASR on the LibriSpeech 960h training set. This grounds the error distribution in real acoustic conditions and complements the synthetic TTS data.

For the cross-domain generalization experiments
(Section~\ref{sec:generalization_domains}), we train a separate ECLM on
Parakeet-TDT-v2~\cite{nvidia2024parakeet} transcriptions of the English subset
of the Granary dataset~\cite{granary2025}, a large-scale open-source collection
of pseudo-labeled speech across 25 European languages sourced from the wild, of
which around 120k hours are English.

\subsection{ASR models}\label{sec:asr}
Table~\ref{tab:asr_models} summarizes the ASR models used in our experiments.
The primary model is a Transformer-based CTC encoder with 255M parameters,
referred to as the \textit{baseline ASR} throughout the paper.
All CTC models (Transformer, Conformer~\cite{conformer}, QuartzNet~\cite{quartznet}) are trained on LibriSpeech 960h with a character vocabulary.
Whisper models~\cite{whisper} and Parakeet-TDT-v2~\cite{nvidia2024parakeet} are pretrained on much larger proprietary audio corpora and use word-piece tokenization.

\begin{table}[h]
\caption{ASR models used in this work.}
\label{tab:asr_models}
\centering
\begin{tabular}{llrl}
\toprule
\textbf{Model} & \textbf{Decoder} & \textbf{Params} & \textbf{Training data} \\
\midrule
Transformer & CTC & 255M & LS 960h \\
Conformer & CTC & 102M & LS 960h \\
QuartzNet & CTC & 7M & LS 960h \\
Whisper-base & Enc-Dec & 74M & proprietary \\
Whisper-small & Enc-Dec & 244M & proprietary \\
Parakeet-TDT-v2 & TDT & 0.6B & Granary \\
\bottomrule
\end{tabular}
\end{table}

The \textit{baseline ASR} (Transformer CTC, 255M) consists of a 1D convolution (kernel 7, stride 3), sinusoidal positional embedding, and 36 pre-LayerNorm transformer blocks (embedding dim 768, 4 heads, MLP dim 3072, dropout 0.1, layer drop 0.1).
The Conformer (102M) uses 16 conformer blocks (embedding dim 512, 4 heads, MLP dim 2048, dropout 0.1).
All acoustic models use 80-channel log-mel filterbanks (25ms window, 10ms stride) with SpecAugment~\cite{spec-aug} (2 frequency masks, max width 30; 10 time masks, max width 50, ratio 0.1).
Models are trained with AdamW (weight decay 1e-6) and gradient clipping of 1.0/0.5/1.0 for Transformer/Conformer/QuartzNet, with peak learning rates of 0.001/0.002/0.0022 and warmup of 64k/10k/64k steps respectively.
Training continues until greedy WER plateaus on \textit{dev-clean} and \textit{dev-other}.

\subsection{Language models}\label{sec:eclm_nlm}
\textbf{ECLM}~~~The ECLM is a Transformer encoder-decoder trained with cross-entropy loss and a character vocabulary, with 16 encoder layers and 4 decoder layers (dropout and layer drop 0.1, sinusoidal positional embedding).
We vary the embedding and MLP dimensions for different model sizes: 512/2048 (69M), 768/3072 (155M), and 1280/6144 (484M), following~\cite{synnaeve2019end}.
The training data is generated from LibriSpeech LM text corpus by different TTS systems but from the \textit{baseline ASR} model.
ECLMs are trained with a dynamic batch size of 160k tokens, AdamW optimizer (weight decay 0.01, gradient clipping 0.1), learning rate warmup of 64k steps, followed by a constant learning rate of 0.001 for 300k steps, then step decay (rate 0.5, step size 200k) until greedy WER plateaus on \textit{dev-clean} and \textit{dev-other}.
The same training hyperparameters are used across all experiments, varying only model size and training data.

\noindent\textbf{Neural LM}~~~The neural LM is a Transformer decoder-only model with 20 layers, trained with cross-entropy loss and a 10K word-piece vocabulary on the LibriSpeech LM text corpus. It shares the same embedding and MLP dimensions with the ECLM.
Training uses AdamW (weight decay $1e{-}8$) with a cosine learning rate schedule (peak 0.001, 500k steps, 16k warmup). Perplexity for 69M, 155M, and 484M LMs is 34.45, 31.49, and 30.9 respectively.
For neural LM rescoring (Figure~\ref{fig:decoding}a), we generate an $n$-best list from CTC beam-search decoding with a 4-gram word-level LM (top-200k words), and rescore using neural LM and CTC ASR scores, similar to~\cite{synnaeve2019end}.

\noindent\textbf{LLM baselines}~~~To benchmark against generic LLMs, we evaluate three instruction-tuned models for ASR error correction:
Llama-3.1-8B-Instruct~\cite{grattafiori2024llama3}, Llama-3.1-70B-Instruct~\cite{grattafiori2024llama3}, and Mistral-7B-Instruct-v0.3~\cite{jiang2023mistral}.
All models are used off-the-shelf without any fine-tuning on ASR data.
We evaluate in both zero-shot and 5-shot settings, where the few-shot examples are drawn from the LibriSpeech dev set. The prompt template is shown in Figure~\ref{fig:llm_prompt}.
Inference is performed with greedy decoding (temperature 0) on a single H100 GPU.

\begin{figure}[h]
\centering
\fbox{\parbox{0.95\columnwidth}{\raggedright
{\small\textbf{System prompt:}}\\[2pt]
{\small\ttfamily You are an ASR error correction system. Correct any transcription errors in the following text. Only fix errors---do not add, remove, or rephrase content. Output only the corrected transcript.}\\[6pt]
{\small\textbf{Few-shot user prompt:}}\\[2pt]
{\small\ttfamily Correct ASR transcription errors. Examples:}\\[2pt]
{\small\ttfamily Input: ``he went too the see yesterday''}\\
{\small\ttfamily Output: ``he went to the sea yesterday''}\\[2pt]
{\small\ttfamily Input: ``their our many problems with this''}\\
{\small\ttfamily Output: ``there are many problems with this''}\\[2pt]
{\small\ttfamily Input: ``the whether is nice today''}\\
{\small\ttfamily Output: ``the weather is nice today''}\\[2pt]
{\small\ttfamily Input: ``she could of done better''}\\
{\small\ttfamily Output: ``she could have done better''}\\[2pt]
{\small\ttfamily Input: ``i need to by some food''}\\
{\small\ttfamily Output: ``i need to buy some food''}\\[2pt]
{\small\ttfamily Input: ``\{hypothesis\}''}\\
{\small\ttfamily Output:}
}}
\caption{Prompt template used for LLM-based ASR error correction (5-shot). The system prompt instructs conservative correction; the few-shot examples demonstrate common ASR error types (homophones, phonetic confusions). Zero-shot uses the system prompt only.}
\label{fig:llm_prompt}
\vspace{-0.4cm}
\end{figure}

\noindent\textbf{Compute}~~~All models are trained on A100 with 80GB memory with 1 node of 8 GPUs for 3-5 days. Some models we train with larger batch size which results in longer training with gradient accumulation.
For data generation by YourTTS model we use CPU inference and it takes 100k CPU hours (as it is not batched version of public code). With RichTTS (116M), we spent 12,800 GPU hours to synthesize 40M audios (roughly 137k hours of data).
Correction-first decoding with beam of 64 is sufficient, while the beam of 10 gives similar results within 0.1\% variation in WER, which takes about 10~mins to finish on 8 A100 GPU for dev sets.

\section{Results}

We organize our experiments to answer five questions: (1)~How does our ECLM compare to neural LM rescoring and LLM-based correction? (2)~What types of errors can ECLMs fix that LLMs cannot? (3)~Does a single ECLM generalize across ASR architectures and domains? (4)~What are the key ingredients for effective synthetic training data? (5)~How does performance scale with model size, data size, and speaker diversity?

We use LibriSpeech (LS) as the primary benchmark throughout, as it is the most widely used evaluation set for ASR, enables direct comparison with prior work, and provides a large text-only LM corpus (800M words) that can be used for synthetic data generation via TTS. Importantly, modern ASR systems already achieve very low WER on LibriSpeech, making it a challenging testbed for error correction: reducing WER from 5\% to 3\% requires precise, character-level fixes rather than coarse linguistic corrections, and any overcorrection risks degrading already-accurate transcriptions. We then evaluate generalization to other architectures and domains in Section~\ref{sec:generalization}.
All WER scores are computed after applying the Whisper text normalizer~\cite{whisper} for consistent evaluation across models.

\subsection{Main results}
\label{sec:librispeech_results}

Table~\ref{tab:main_res} compares neural LM rescoring, LLM-based correction, and
our ECLM on LibriSpeech test set.
\begin{table}[h]
  \caption{LibriSpeech test sets WER (\%) comparing neural LM rescoring, LLM-based correction, and our ECLM. Latency measured on a single H100 GPU.}
  \label{tab:main_res}
  \centering
  \begin{tabular}{lccccc}
\toprule
\textbf{Model} & \textbf{Params} & \textbf{LS-clean} & \textbf{LS-other} &
\textbf{Latency (ms)} & \textbf{Halluc. (\%)} \\
\midrule
\textit{baseline ASR} & -- & 2.2 & 5.3 & -- & -- \\
\midrule
\multicolumn{6}{l}{\textit{Neural LM}} \\
~~+ NeLM rescoring & 0.5B & 2.0 & 4.1 & -- & -- \\
\midrule
\multicolumn{6}{l}{\textit{Generic LLMs}} \\
~~+ Mistral & \multirow{2}{*}{7B} & & & & \\
\hspace{16pt} 0-shot & & 32.0 & 37.0 & 579 & 11.5 \\
\hspace{16pt} 5-shot & & 20.0 & 25.8 & 774 & 6.6 \\
~~+ Llama & \multirow{2}{*}{8B} & & & & \\
\hspace{16pt} 0-shot & & 16.7 & 23.3 & 295 & 4.9 \\
\hspace{16pt} 5-shot & & 7.1 & 13.4 & 243 & 3.1 \\
~~+ Llama & \multirow{2}{*}{70B} & & & & \\
\hspace{16pt} 0-shot & & 8.8 & 13.0 & 1202 & 4.6 \\
\hspace{16pt} 5-shot & & 19.3 & 19.5 & 1767 & 4.9 \\
\midrule
\multicolumn{6}{l}{\textit{Specialized LMs}} \\
~~+ ECLM (greedy) & 0.5B & 2.0 & 4.1 & 44 & 0 \\
~~+ ECLM (corr.-first) & 0.5B & \textbf{1.6} & \textbf{3.6} & 457 & 0 \\
\bottomrule
\end{tabular}
\vspace{-0.2cm}
\end{table}

\noindent\textbf{ECLM vs.\ neural LM rescoring}
Our ECLM with greedy decoding achieves comparable WER to neural LM rescoring, despite not requiring beam search from the ASR.
With correction-first decoding, our model significantly outperforms neural LM rescoring (1.6\%/3.6\% vs.\ 2.0\%/4.1\%), achieving 20\% and 12\% relative WER reduction on LS-clean and LS-other respectively.

\noindent\textbf{ECLM vs.\ generic LLMs}
We evaluate LLMs using the prompt template shown in Figure~\ref{fig:llm_prompt}.
All LLMs degrade the baseline ASR, even the best result---Llama-70B zero-shot (8.8\%/13.0\%)---which has 140$\times$ more parameters than our ECLM.
Scaling from 8B to 70B yields only modest WER improvement at 4--6$\times$ higher latency (1202~ms vs.\ 243--295~ms), suggesting that LLMs do not scale efficiently for error correction in the low-WER regime.
Few-shot prompting helps smaller models (Llama-8B 5-shot: 7.1\%/13.4\%) but hurts the 70B model (19.3\%/19.5\%), likely due to over-reliance on example patterns.
\textit{We also fine-tuned Mistral-7B-v0.1 with both LoRA and full fine-tuning on our
training data, where full fine-tuning achieving 2.0\%/4.8\% WER---still worse
than our ECLM with greedy (2.0\%/4.1\%) and 
correction-first decoding (1.6\%/3.6\%), confirming that architecture matters
more than simply adapting a generic LLM.}
Our ECLM runs at 44~ms---5--40$\times$ faster than LLMs.
More critically, all LLMs exhibit high hallucination rates (3--12\%), whereas our ECLM produces zero hallucinations.
We define hallucination rate as the percentage of words in the model's output that appear in neither the ASR hypothesis nor the reference transcript, measuring words fabricated without acoustic or textual support.

\begin{figure*}[h]
\centering
\includegraphics[width=\textwidth]{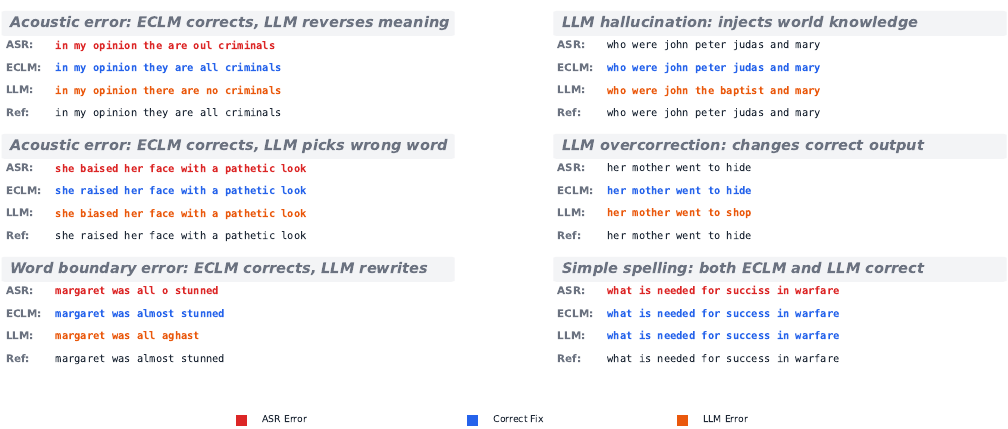}
\caption{Qualitative comparison on dev-other. \textcolor{red}{Red}: ASR errors, \textcolor{blue}{blue}: correct ECLM fixes, \textcolor{orange}{orange}: LLM errors.}
\label{tab:error_examples}
\vspace{-0.2cm}
\end{figure*}

\subsection{Qualitative analysis: why do LLMs struggle?}
\label{sec:llm_comparison}

To understand why LLMs underperform despite their strong language modeling capabilities, we examine error types qualitatively.
We categorize ASR errors into: (1)~\textit{acoustic errors}---phonetic confusions and homophones where the incorrect word sounds similar to the correct one; and (2)~\textit{linguistic errors}---grammar and word choice issues unrelated to sound.

Figure~\ref{tab:error_examples} shows representative examples.
LLMs struggle with acoustic errors: ``baised''$\rightarrow$``raised'' requires knowledge of ASR error patterns, but the LLM picks the linguistically plausible ``biased'' instead. Similarly, the LLM rewrites ``all o stunned'' as ``all aghast'' rather than recovering the correct word boundary ``almost stunned''.
More dangerously, LLMs can reverse meaning entirely---correcting ``the are oul criminals'' to ``there are no criminals'' instead of ``they are all criminals''.
Our ECLM, trained on synthetic and real ASR errors that capture actual error distributions, learns these acoustic confusion patterns and makes precise character-level corrections.

Most critically, LLMs hallucinate on already-correct ASR output. They inject world knowledge (``john peter judas''$\rightarrow$``john the baptist''), or substitute plausible but incorrect words (``hide''$\rightarrow$``shop'').
In the low-WER regime where remaining errors are subtle phonetic confusions, LLMs over-correct while our ECLM makes conservative, targeted fixes.

\subsection{Generalization across architectures and domains}
\label{sec:generalization}

\subsubsection{Generalization across ASR architectures}

\begin{table}[h]
  \caption{LibriSpeech test WER (\%) across ASR architectures. The same ECLM and neural LM from Table~\ref{tab:main_res} are used without retraining.}
  \label{tab:universality}
  \centering
  \begin{tabular}{lrcc}
\toprule
\textbf{Model} & \textbf{Params} & \textbf{LS-clean} & \textbf{LS-other}\\
\midrule
QuartzNet-CTC & \multirow{4}{*}{7M} & 6.5 & 16.9 \\
\hspace{4pt}+ neLM (rescoring) & & 2.9 & 8.6 \\
\hspace{4pt}+ ECLM (greedy) & & 2.8 & 8.3 \\
\hspace{4pt}+ ECLM (corr.-first) & & 2.3 & 7.1\\
\midrule
Conformer-CTC & \multirow{4}{*}{102M} & 2.6 & 5.6 \\
\hspace{4pt}+ neLM (rescoring) & & 2.2 & 4.2 \\
\hspace{4pt}+ ECLM (greedy) & & 2.2 & 4.1 \\
\hspace{4pt}+ ECLM (corr.-first) & & 1.7 & 3.6 \\
\midrule
Whisper-base & \multirow{2}{*}{74M} & 4.5 & 10.9\\
\hspace{4pt}+ ECLM (greedy) & & 3.1 & 7.9\\
\midrule
Whisper-small & \multirow{2}{*}{244M} & 3.3 & 7.6\\
\hspace{4pt}+ ECLM (greedy) & & 2.7 & 6.1\\
\bottomrule
\end{tabular}
\end{table}

To test whether a single ECLM trained on noise from the \textit{baseline ASR}
transfers to other ASR architectures and sizes, we apply the same ECLM (0.5B) to
QuartzNet (7M), Conformer (102M), and Whisper (base/small) without any retraining.
Table~\ref{tab:universality} shows that the ECLM consistently improves all ASR models, with correction-first decoding outperforming neural LM rescoring across architectures.
Notably, the ECLM improves Whisper---an encoder-decoder model trained on proprietary data with word-piece tokenization---despite being trained only on CTC-based ASR errors with character tokenization, demonstrating that error patterns generalize across architectures.


\subsubsection{Generalization across domains and datasets}
\label{sec:generalization_domains}

To evaluate whether the proposed error correction approach generalizes beyond LibriSpeech, we train an ECLM (0.5B) on Parakeet-TDT-v2~\cite{nvidia2024parakeet} transcriptions of the English subset of the Granary dataset~\cite{granary2025}.
Both Parakeet hypotheses and Granary labels are lowercased and stripped of punctuation (except hyphens) before training and evaluation, to prevent the ECLM from spending capacity on trivial casing and punctuation differences rather than learning meaningful error corrections.
We then apply this ECLM to correct Parakeet outputs across eight diverse
datasets spanning two categories: conversational and noisy speech
(CHiME-6~\cite{watanabe2020chime}, CallHome~\cite{cieri2004fisher},
Switchboard~\cite{cieri2004fisher}, CommonVoice~\cite{ardila2019common}), and
read or prepared speech with already-low baseline WER (LibriSpeech~\cite{panayotov2015librispeech},
VoxPopuli~\cite{wang2021voxpopuli}, TED-LIUM~3~\cite{hernandez2018ted}).
This setting reflects real-world deployment, where a single correction model must generalize across domains, acoustic conditions, and speaker populations without per-dataset adaptation.

\begin{table}[h]
\caption{WER (\%) for Parakeet-TDT-v2 corrected by our ECLM (0.5B) across diverse datasets.}
\label{tab:parakeet_generalization}
\centering
\begin{tabular}{lccc}
\toprule
\textbf{Dataset} & \textbf{Parakeet} & \textbf{+ ECLM (greedy)} & \textbf{Rel. $\downarrow$} \\
\midrule
\multicolumn{4}{l}{\textit{Conversational / noisy}} \\
CHiME-6 & 34.0 & 25.8 & 24\% \\
CallHome & 15.5 & 13.5 & 13\% \\
Switchboard & 11.1 & 10.3 & 7\% \\
CommonVoice (en) & 10.1 & 10.0 & 2\% \\
\midrule
\multicolumn{4}{l}{\textit{Prepared / low-WER}} \\
VoxPopuli (en) & 6.5 & 6.4 & 1\% \\
LS-clean & 2.1 & 2.1 & 0\% \\
LS-other & 4.0 & 4.0 & 0\% \\
TED-LIUM 3 & 3.9 & 3.9 & 0\% \\
\midrule
\textit{Average} & 10.9 & 9.5 & 13\% \\
\bottomrule
\end{tabular}

\end{table}

Table~\ref{tab:parakeet_generalization} reflects several key results.
On conversational and noisy datasets (CHiME-6, CallHome, Switchboard), where Parakeet produces frequent errors due to overlapping speech, disfluencies, and background noise, the ECLM provides substantial improvements: 24\% relative reduction on CHiME-6, 13\% on CallHome, and 7\% on Switchboard.
On prepared speech datasets (LibriSpeech, TED-LIUM, VoxPopuli), where Parakeet already performs well, the ECLM preserves performance without degradation.

The variation in gains across datasets is driven by multiple factors beyond ASR accuracy alone.
First, the Granary training labels are pseudo-labeled by Whisper, so the ECLM's correction ceiling is bounded by the quality of these labels---on domains where Whisper itself is less accurate, the ECLM has less room to improve.
Second, the domain distribution matters: conversational datasets are acoustically closer to the diverse conditions in Granary, while prepared speech domains like LibriSpeech and TED-LIUM are underrepresented.
Third, when the input hypothesis is already correct, the ECLM learns to pass it through unchanged, avoiding the overcorrection problem that plagues LLMs.
On average, the ECLM reduces WER from 10.9\% to 9.5\% (13\% relative), demonstrating that a single ECLM can serve as a universal post-processor across ASR architectures, domains, and acoustic conditions.

\subsection{Ablations}

As argued in Section~\ref{sec:methods}, the effectiveness of ECLMs depends critically on matching the training error distribution to real ASR errors.
We systematically ablate three axes: the composition of training noise (Section~\ref{sec:noise_effect}), the role of TTS system choice and synthetic vs.\ real data (Section~\ref{sec:tts_analysis}), and scaling behavior along model size, data size, and speaker diversity (Section~\ref{sec:scaling}).
All ablations are evaluated on the LibriSpeech dev sets.

\subsubsection{Noise distribution in training data}
\label{sec:noise_effect}

Table~\ref{tab:noise_effect} traces the data construction path to our best results, showing the incremental effect of each strategy.
Starting from 40M $\{(\hat{y}, y)\}$ pairs generated by YourTTS, the baseline ECLM improves \textit{dev-other} but slightly degrades \textit{dev-clean} (row~2), suggesting the synthetic errors are too clean for the model to learn meaningful corrections.
We progressively diversify the error distribution through three strategies, each carefully chosen to avoid shifting the training distribution too far from real ASR errors:
(1)~random character substitution at rate $s{=}10\%$ (row~4);
(2)~frequency masking of the spectrogram before ASR inference, similar to SpecAugment~\cite{spec-aug} (row~5);
(3)~mixing in 10\% real ASR hypotheses from the LibriSpeech 960h training set to ground the distribution in real acoustic conditions (row~7).
Additionally, combining data from two TTS systems (YourTTS + RichTTS) outperforms either system alone (row~6), as different TTS models produce complementary error patterns.
Combining all strategies yields the best results (row~8).

\begin{table}[h]
  \caption{LibriSpeech dev WER (\%) under different training data compositions. $s$: character substitution rate; FM: frequency masking.}
  \label{tab:noise_effect}
  \centering
  \begin{tabular}{lcccc}
\toprule
 & \multicolumn{2}{c}{\textbf{Greedy}} & \multicolumn{2}{c}{\textbf{Corr.-first}} \\
\cmidrule(lr){2-3} \cmidrule(lr){4-5}
\textbf{Model} & \textbf{clean} & \textbf{other} & \textbf{clean} & \textbf{other} \\
\midrule
\textit{baseline ASR} & 2.1 & 5.5 & -- & -- \\
\midrule
\hspace{4pt}+ ECLM, Tacotron + $s{=}10\%$ & 4.1 & 6.6 & 2.0 & 4.8 \\
\midrule
\hspace{4pt}+ ECLM (YourTTS) & 2.3 & 4.9 & 1.6 & 4.0 \\
\hspace{8pt}+ $s{=}10\%$ & 2.2 & 4.6 & 1.5 & 3.8 \\
\hspace{12pt}+ FM & 2.3 & 4.6 & 1.6 & 3.7 \\
\hspace{12pt}+ RichTTS & 2.1 & 4.2 & 1.5 & 3.7 \\
\hspace{12pt}+ FM + real & 2.0 & 4.3 & 1.5 & 3.8 \\
\hspace{12pt}+ RichTTS + FM + real & 1.9 & 3.9 & 1.5 & 3.4 \\
\midrule
\hspace{4pt}+ ECLM, RichTTS + $s{=}10\%$ & 2.2 & 4.3 & 1.6 & 3.6 \\
\bottomrule
\end{tabular}

\end{table}

\subsubsection{Synthetic data analysis}
\label{sec:tts_analysis}

\noindent\textbf{TTS audio vs.\ real audio}~~~To assess whether synthetic data can substitute for real labeled audio, we compare ECLMs trained on TTS-generated hypotheses vs.\ hypotheses from $\sim$60k hours of real speech from Libriheavy~\cite{kang2023libriheavy} (400M words), both decoded by the \textit{baseline ASR}. For a fair comparison, we generate TTS data with RichTTS for the same sentences.
As shown in Table~\ref{tab:real-synt}, the gap is small: ECLM-synthetic is slightly worse on \textit{dev-clean} (1.7\% vs.\ 1.5\%) but outperforms ECLM-real on \textit{dev-other} (3.7\% vs.\ 4.1\%), suggesting that TTS-generated diversity compensates for the lack of real acoustic conditions. This validates the TTS-based pipeline as a scalable alternative to collecting labeled audio.

\begin{table}[h]
\caption{LibriSpeech dev WER (\%) for ECLMs trained on synthetic vs.\ real audio.}
\label{tab:real-synt}
\centering
\begin{tabular}{lcccc}
\toprule
 & \multicolumn{2}{c}{\textbf{Greedy}} & \multicolumn{2}{c}{\textbf{Corr.-first}} \\
\cmidrule(lr){2-3} \cmidrule(lr){4-5}
\textbf{Model} & \textbf{clean} & \textbf{other} & \textbf{clean} & \textbf{other} \\
\midrule
\textit{baseline ASR} & 2.1 & 5.5 & -- & -- \\
\midrule
\hspace{4pt}+ ECLM-synthetic & 2.4 & 4.5 & 1.7 & 3.7 \\
\hspace{4pt}+ ECLM-real & 2.0 & 4.6 & 1.5 & 4.1 \\
\bottomrule
\end{tabular}
\end{table}

\noindent\textbf{TTS quality vs.\ error diversity}~~~We evaluate the audio quality of each TTS system by measuring ASR WER on the synthesized speech (Table~\ref{tab:tts_eval}). Tacotron produces the cleanest audio (6\% WER with Whisper-small), followed by YourTTS (10\%) and RichTTS (12\%).
However, when training ECLMs on data from each system separately, Tacotron performs worst for error correction despite its higher audio quality (Table~\ref{tab:noise_effect}, row~2 vs.\ rows~3--4).
This is because cleaner audio leads to fewer ASR errors in the training pairs, leaving the ECLM with mostly identity mappings and insufficient signal to learn meaningful corrections.
Additionally, Tacotron is single-speaker, further limiting the diversity of error patterns compared to the multi-speaker YourTTS and RichTTS.
This finding is counterintuitive but consistent with our central thesis: what matters for ECLM training is not TTS fidelity, but the richness and diversity of the resulting error distribution.

\begin{table}[h]
\caption{LibriSpeech dev WER (\%) on TTS-generated audio and resulting ECLM performance (greedy) when trained on each TTS system.}
\label{tab:tts_eval}
\centering
\begin{tabular}{lcccccc}
\toprule
 & \multicolumn{2}{c}{\textbf{Whisper-small}} & \multicolumn{2}{c}{\textbf{Baseline ASR}} & \multicolumn{2}{c}{\textbf{ECLM (greedy)}} \\
\cmidrule(lr){2-3} \cmidrule(lr){4-5} \cmidrule(lr){6-7}
\textbf{TTS System} & \textbf{clean} & \textbf{other} & \textbf{clean} & \textbf{other} & \textbf{clean} & \textbf{other} \\
\midrule
Tacotron & 5.9 & 5.2 & 6.8 & 6.2 & 4.1 & 6.6 \\
YourTTS & 8.2 & 9.8 & 8.6 & 10.5 & 2.3 & 4.9 \\
RichTTS & 7.7 & 11.9 & 9.5 & 16.7 & 2.2 & 4.3 \\
\bottomrule
\end{tabular}

\end{table}

\subsubsection{Scalability analysis}
\label{sec:scaling}

\begin{figure*}[h]
\centering
\includegraphics[width=\textwidth]{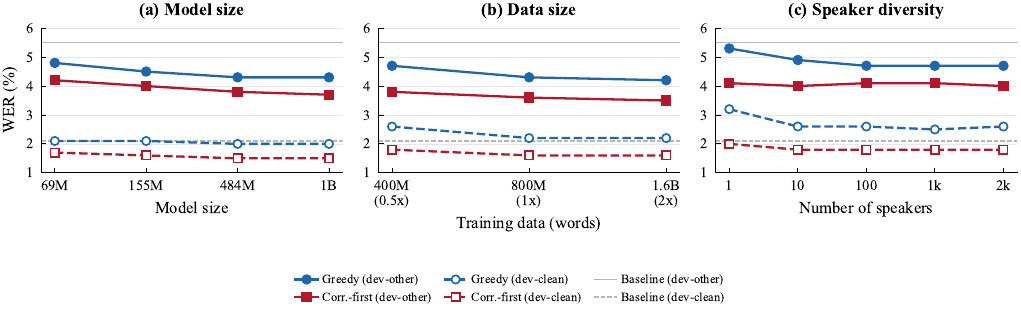}
\caption{LibriSpeech dev WER (\%) as a function of model size, training data size, and speaker diversity. Solid lines: dev-other; dashed lines: dev-clean.}
\label{fig:scaling}
\end{figure*}

A key practical question is whether ECLMs follow predictable scaling behavior.
We find that ECLMs benefit consistently from scaling along three axes (Figure~\ref{fig:scaling}).

\noindent\textbf{Model size}~~~WER decreases consistently as model size increases from 69M to 1B for both greedy and correction-first decoding (Figure~\ref{fig:scaling}a), with the largest gains between 69M and 484M.

\noindent\textbf{Text corpus size}~~~We construct three corpora by subsampling or augmenting the LibriSpeech LM corpus: 0.5x (400M words), 1x (800M words), and 2x (1.6B words, augmented with Gutenberg text~\cite{xu20b_interspeech}).
Larger corpora yield lower WER (Figure~\ref{fig:scaling}b), confirming that ECLMs benefit from greater textual diversity in training.

\noindent\textbf{Speaker diversity}~~~We train five ECLMs (155M) on data synthesized from 1, 10, 100, 1k, and 2k speakers.
More speakers reduce WER, particularly for greedy decoding, by increasing the variability of ASR error patterns in the training data (Figure~\ref{fig:scaling}c).
Gains saturate around 100 speakers on LibriSpeech dev sets; we use all available speakers for final experiments.

\subsection{Comparison with prior work}
\label{sec:sota}

A natural question is how our approach compares to systems that achieve low WER through fundamentally different strategies: larger architectures, self-supervised pretraining on unlabeled audio, or massive supervised training corpora.
Table~\ref{tab:compare} places our best system in context.
Following prior work showing that TTS data can improve ASR~\cite{hu2021synt}, we also train a Transformer ASR variant with a mixture of LibriSpeech and TTS audio, denoted \textit{baseline ASR (LS+TTS)}.
This stronger baseline with correction-first decoding achieves 1.5\% and 3.3\% WER on test-clean and test-other, respectively.
Without using any external audio data, our results match wav2vec 2.0 and HuBERT models that leverage 60k hours of unlabeled audio for self-supervised pretraining, and outperform Zipformer~\cite{yao2024zipformer}, the strongest specialized model in the no-external-audio category.
Our results are also competitive with state-of-the-art models from the Open ASR Leaderboard, including Granite Speech 3.3-8B~\cite{saon2025granite} (1.4\%/2.9\%), Canary-Qwen-2.5B~\cite{nvidia2025canary} (1.6\%/3.1\%), and Qwen3-ASR-1.7B~\cite{shi2026qwen3} (1.6\%/3.4\%)---all of which use significantly more parameters and large-scale training data.
This suggests that error correction is a complementary and cost-effective path to low WER, orthogonal to scaling acoustic models or training data.
\begin{table}[h]
  \caption{LibriSpeech test WER (\%) comparison with prior work.}
  \label{tab:compare}
  \centering
  \begin{tabular}{lccc}
\toprule
\textbf{Model} & \textbf{Audio Data} & \textbf{LS-clean} & \textbf{LS-other} \\
\midrule
\multicolumn{4}{l}{\textit{No external audio data}} \\
Transformer~\cite{synnaeve2019end} & LS-960h & 2.3 & 5.2 \\
Context-Net (L)~\cite{contextnet} & LS-960h & 1.9 & 4.1 \\
Conformer (Transducer)~\cite{conformer} & LS-960h & 1.9 & 3.9 \\
ASAPP-ASR~\cite{asapp} & LS-960h & 1.8 & 4.5 \\
E-branchformer + ILME~\cite{ebranchformer} & LS-960h & 1.8 & 3.7 \\
Zipformer~\cite{yao2024zipformer} & LS-960h & 1.6 & 3.6 \\
SYNT++~\cite{hu2021synt} & LS-960h & 2.4 & 6.3 \\
LAS + SC + LM~\cite{guo2019spelling} & LS-960h & 4.3 & -- \\
\midrule
\multicolumn{4}{l}{\textit{Self-supervised pretraining (LL-60k)}} \\
wav2vec 2.0-Large~\cite{baevski2020wav2vec} & LL-60k & 1.8 & 3.3 \\
data2vec 2.0~\cite{baevski2023data2vec} & LL-60k & 1.7 & 3.0 \\
HuBERT-Large~\cite{hsu2021hubert} & LL-60k & 1.9 & 3.3 \\
HuBERT-XL~\cite{hsu2021hubert} & LL-60k & 1.8 & 2.9 \\
Conformer XXL~\cite{zhang2020pushing} & LL-60k & 1.5 & 3.1 \\
\midrule
\multicolumn{4}{l}{\textit{Large-scale supervised / Speech LLMs}} \\
Whisper large-v3~\cite{whisper} & 5M hrs & 2.0 & 3.7 \\
Parakeet-TDT-v2~\cite{nvidia2024parakeet} & 120k hrs & 1.7 & 3.2 \\
Canary-Qwen-2.5B~\cite{nvidia2025canary} & 234k hrs & 1.6 & 3.1 \\
Granite Speech 3.3-8B~\cite{saon2025granite} & 76k hrs & 1.4 & 2.9 \\
Qwen3-ASR-1.7B~\cite{shi2026qwen3} & 40M hrs + & 1.6 & 3.4 \\
\midrule
\textit{baseline ASR (LS+TTS)} + corr.-first & LS-960h & 1.5 & 3.3 \\
\bottomrule
\end{tabular}

\end{table}

\section{Conclusion}
We revisit ASR error correction with specialized language models.
Our experiments demonstrate that ECLMs---compact seq2seq models trained on a
mixture of ASR errors from real and synthetic audio---outperform both neural LM rescoring and generic LLMs,
including fine-tuned variants, while being 15$\times$ smaller, 5--40$\times$
faster, and free of hallucination.

The central insight is that matching the training error distribution to real ASR mistakes matters more than model scale.
Through systematic ablations, we show that multi-speaker TTS, noise augmentation, mixing synthetic and real data, and---counterintuitively---noisier TTS systems all contribute to a more representative error distribution.

We propose correction-first decoding, which reverses the conventional LM rescoring pipeline and outperforms it at comparable computational cost.
A single ECLM generalizes across ASR architectures (QuartzNet, Conformer, Whisper, Parakeet) spanning CTC, seq2seq, and transducer paradigms, and across domains, reducing WER by 13\% relative on average across eight diverse datasets spanning conversational and prepared speech.

Our results suggest that in the era of large language models, \textit{compact specialized models} that capture \textit{task-specific distributions} remain a compelling alternative---particularly in the low-error regime where LLMs \textit{overcorrect} and \textit{hallucinate}.

\bibliographystyle{plainnat}
\bibliography{sample}

\end{document}